\documentclass{article}

\usepackage[final]{neurips_2021}

\usepackage[utf8]{inputenc} 
\usepackage[T1]{fontenc}    
\usepackage{hyperref}       
\usepackage{url}            
\usepackage{booktabs}       
\usepackage{amsfonts}       
\usepackage{nicefrac}       
\usepackage{microtype}      
\usepackage{xcolor}         
\usepackage{amsmath}
\usepackage{graphicx}
\usepackage{subcaption}
\usepackage{placeins}

\newcommand{\rand}{\text{rand}}

\title{Causal policy ranking}

\author{Daniel McNamee, Hana Chockler \\
  causaLens \\
  London, UK \\
  \texttt{dan@causalens.com} \\
}

\begin{document}

\maketitle

\begin{abstract}
Policies trained via reinforcement learning (RL) are often very complex even for simple tasks. In an episode with $n$ time steps, a policy will make $n$ decisions on actions to take, many of which may appear non-intuitive to the observer. Moreover, it is not clear which of these decisions directly contribute towards achieving the reward and how significant is their contribution. Given a trained policy, we propose a black-box method based on counterfactual reasoning that estimates the causal effect that these decisions have on reward attainment and ranks the decisions according to this estimate. In this preliminary work, we compare our measure against an alternative, non-causal, ranking procedure, highlight the benefits of causality-based policy ranking, and discuss potential future work integrating causal algorithms into the interpretation of RL agent policies.
\end{abstract}

\section{Introduction}

Reinforcement learning is a powerful 
method for training policies that
complete tasks in complex environments via sequential action selection \citep{Sutton2018}.
The policies produced are optimised to maximise 
the expected cumulative reward provided by the environment. While
reward maximization is clearly an important goal, 
this single measure may not reflect other objectives that an engineer or scientist may desire in training RL agents. Focusing solely on 
performance risks overlooking the demand for
models that are easier to analyse, 
predict and interpret~\citep{lewis2020deep}.
Our hypothesis is that many trained policies
are \emph{needlessly complex}, i.e., that there
exist alternative policies that
perform just as well or nearly as well
but that are significantly simpler.

The starting point for our definition of ``simplicity'' is
the assumption that there exists a way to make a ``simple choice''
based on randomly selecting an action from the available repertoire
in an environment. We argue that this may be the case
for many environments in which RL is applied. That is, a agent may be able to simplify its policy by randomizing its action selection in some states without a drastic drop in expected reward obtained. The tension between performance and simplicity is central to the field of explainable AI (XAI), and 
machine learning as a whole~\citep{gunning2019darpa}.
The key contribution of this paper is a novel causality-based method for
simplifying policies while minimizing the compromise with respect to performance,
hence addressing one of the main hurdles for wide adoption
of RL: the high complexity of trained policies.
%

We introduce an algorithm for
\emph{ranking the importance of the decisions} that a policy makes,
by scoring the actions it selects. The rank reflects
the impact that replacing the policy's chosen action by
a randomly selected action has on the reward outcome.
We develop rankings based on sample-based estimates since it is intractable to compute this ranking precisely, due to the
high complexity and the stochasticity of the environment and the policy, 
complex causal interactions between actions and their outcomes,
and the sheer size of the problem. 

Ranking policy decisions according to their importance was recently introduced
by~\citep{PCSK20}, who use spectrum-based fault localization techniques
to approximate the contribution of decisions to reward attainment.
%
Our algorithm uses \emph{causal effects} techniques~\citep{Pearl2009} to compute 
the ranking of policy decisions and can be applied to black-box policies, making no assumptions
about the policy's training or representation. 
We use the same proxy measure for evaluating the quality of our ranking
as~\citep{PCSK20}: we construct new, simpler policies (``pruned policies'') that only use the top-ranked decisions,
without retraining, and compare the reward achieved by these policies with the original policy's one. 
Experiments with agents
for MiniGrid~\citep{gym_minigrid}
demonstrate that pruned policies can maintain high performance and also that performance monotonically approaches that of the original policy as more highly ranked decisions are included from the original policy.
As pruned policies are much easier to understand than the original policies,
we consider this a potentially useful method in the context of explainable RL. 
As pruning a given policy does
not require re-training, the procedure is relatively lightweight.
Furthermore,
the ranking of states by itself provides an important insight into the importance
of particular decisions for the performance of the policy overall.

Our method demonstrates that causal counterfactual reasoning \citep{Hum39,Lew73} is applicable to model-free RL
and opens the door for other causality-inspired methods that can further improve the
interpretability and explainability of RL policies, as well as simplify them. For example, previous work has sought to identify simplified policies via regularization. Specifically, a penalty term measuring the KL-divergence between a trained policy and a random baseline is traded off against the expected cumulative reward objective \citep{Tishby2010,Todorov2009}. 
The main difference with our work is that we describe an offline technique that is based on causal reasoning.
We envision an avenue for a potential integration of our approach with the policy regularization approach whereby causal reasoning may be leveraged in identifying a regularized policy.

\section{Background and Definitions}
\label{sec:background}

\subsection{Reinforcement Learning)}\label{sec:rl}

We use a standard reinforcement learning (RL) setup and assume that the reader is familiar with the basic concepts.
An \emph{environment} in RL is defined as a Markov decision process (MDP) with components $\left\{ S, A, P, R, \gamma\right\}$, where $S$ is the set of states $s$, $A$ is the set of actions $a$, $P$ is the transition function, $R$ is the reward function, and $\gamma$ is the discount factor. An agent seeks to learn
a policy $\pi:S\rightarrow A$ that maximises the total discounted reward. Starting from the initial state $s_0$ and given the policy $\pi$, the state-value function is the expected future discounted reward as follows:
\begin{equation}\label{eq:rlvalue}
V_{\pi}(s_0) = \mathbb{E}\left(\sum_{t=0}^{\infty}\gamma^tR(s_t,\pi(s_t),s_{t+1})\right).
\end{equation}
A policy $\pi: S \rightarrow A$ maps states to the actions taken in these states and may be  stochastic. We treat the policy as a black box, and hence make no further assumptions about $\pi$.

\subsection{Credit assignment metrics}

\subsubsection{Causal counterfactual reasoning}

Causal counterfactual reasoning was introduced by \citet{Hum39}, who was the first to identify causation with counterfactual dependence. The counterfactual interpretation of causality
was extended and formalized in~\citet{Lew73}. Essentially, an event $A$ is a cause of an event $B$
if $A$ happened before $B$ and in a possible world where $A$ did not happen, $B$ did not happen either.
In this work, we adapt these concepts to RL, where an event $A$ is a decision of a given RL policy in a given state and $B$ is the success in achieving the reward.

Based on this conceptual framework, we evaluate the significance of a particular action according to its \emph{causal contribution to reward maximization}. The causal contribution measured by
the \emph{causal effect} $C(s,a)$ of a state-action pair, which we define as the difference in expected cumulative reward obtained with respect to alternative (counterfactual) actions $a'$.
Specifically, suppose an agent selects action $a_t$ in state $s_t$ on timestep $t$, then
\begin{equation}\label{eq:causal_effect}
 C(s,a)  = V_{\pi, a_t}(s_0) - \mathbb{E}_{a'_t\sim \pi_\rand(\cdot|s_t)}\left[ V_{\pi, a'_t}(s_0) \right],
\end{equation}
where $V_{\pi, a_t}(s_0)$ is the value function associated with policy $\pi$ except at time $t$ when the action $a_t$ is selected, $\pi_\rand$ denotes the random policy, and $a'_t\sim \pi_\rand(\cdot|s_t)$ implies that action $a'_t$ is sampled from the policy $\pi_\rand(\cdot|s_t)$ at state $s_t$.

\subsubsection{Spectrum-based fault localization (SBFL)}
\label{sec:SFL}

As a comparative ranking measure, we consider spectrum-based fault localization (SBFL).
SBFL techniques~\citep{naish2011model}
have been widely used as an efficient approach to aid in locating the
causes of failures in sequential programs. 
SBFL techniques rank program elements (say program statements)
based on their \emph{suspiciousness scores}, which are computed using correlation-based
measures. Intuitively, a program element is more suspicious if it appears in failed
executions more frequently than in correct executions, and the exact formulas differ between
the measures. Diagnosis of the faulty program can then be conducted by
manually examining the ranked list of elements
in descending order of their suspiciousness until the cause of the fault is found. It has
been shown that SBFL techniques perform well in complex programs \citep{abreu2009practical}. 

The SBFL procedure first executes the program under test using a set of inputs called the \emph{test suite}.
It records the program executions together with a set of Boolean flags that indicate whether a particular program element was executed by the current test.
The task of a fault localization tool is to compute a ranking of the program elements based on the values of these flags. Following the notation from~\citep{naish2011model}, the suspiciousness score of each program statement $s$ is calculated
from a set of parameters $\langle a^s_\mathit{ep}, a^s_\mathit{ef}, a^s_\mathit{np}, a^s_\mathit{nf} \rangle$ that
give the number of times the statement $s$ is executed ($e$) or not executed~($n$) on passing ($p$)
and on failing ($f$) tests. For instance, $a^s_\mathit{ep}$ is the number of tests that passed and executed $s$. There is a number of SBFL measures, based on different formulae that use these scores; some of the most popular 
include~\citep{ochiai1957zoogeographic,zoltar,jones2005empirical,wong2007effective}.

\begin{figure}
        \centering
        \includegraphics[width=0.43\linewidth]{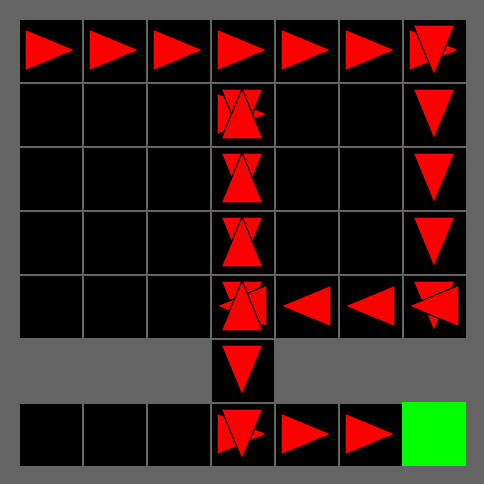}\hfill%
        \includegraphics[width=0.43\linewidth]{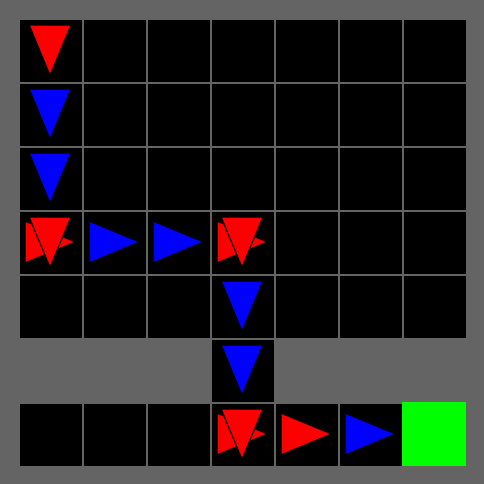}
        \caption{Traces of executions with the original policy (left) and a \emph{pruned policy} (right). States in which 
        we take a random action are in blue. Both policies succeed, but pruning unimportant actions simplifies the policy.}
        \label{fig_prune}
\end{figure}

Recent work on the application of SBFL to RL demonstrated that
SBFL techniques perform well on ranking policy decisions according to their suspiciousness score~\citep{PCSK20}. As a proxy for the quality of the ranking, the SBFL ranking was used to construct simpler
\emph{pruned}
policies by only taking the high-ranked actions from the original policy and substituting low-ranked actions with a randomly generated action (see Figure~\ref{fig_prune} for a demonstration).
Here, we consider a similar approach with causal counterfactual ranking.

\subsection{Ranking policy decisions} 
\label{sec:rankings}

For SBFL ranking, we 
first create the test suite of mutant 
executions $\mathcal{T}(\pi)$ as described above.
We call the set of all abstract states encountered when generating the test suite
$S_{\mathcal{T}} \subseteq \hat S$; these are the states to which we 
assign scores. Any unvisited state is given the lowest possible score by default.

Similarly to SBFL for bug localisation, for each state $s\in S_{\mathcal{T}}$ we
calculate a vector $\langle a^s_\mathit{ep}, a^s_\mathit{ef}, a^s_\mathit{np}, 
a^s_\mathit{nf} \rangle$. We use this vector to track 
the number of times that $s$ was unmutated ($e$) or mutated ($n$)
on passing ($p$) and on failing ($f$) mutant executions, and we do not update
these scores based on executions in which the state was not visited.
In other words, the vector keeps track of
success and failure of mutant executions based on whether an execution took a random action in $s$ or not.
For example, $a^s_{ep}$ is the number of passing executions that took the action $\pi(s)$ 
in the state $s$, and $a^s_{nf}$ is the number of failing executions that took a random action
in the state~$s$. Once we have constructed the vector $\langle a^s_\mathit{ep}, a^s_\mathit{ef}, a^s_\mathit{np}, a^s_\mathit{nf} \rangle$
for every (abstract) state, we apply the SBFL measures discussed in Sec.~\ref{sec:SFL} to rank the states in $\pi$.
This ranking is denoted by $\mathit{rank}: \hat S \rightarrow \{1,\ldots,|\hat S|\}$.


We follow a similar procedure for the causal counterfactual ranking except we replace the SBFL measures with the causal effect estimate (Eqn.~\ref{eq:causal_effect}).

\begin{figure}
  \centering
  \includegraphics[width=0.6\textwidth]{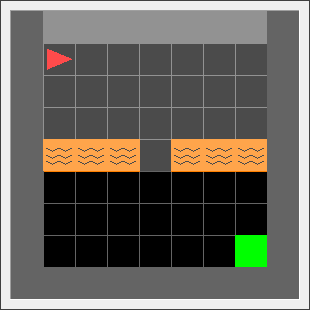}
  \captionof{figure}{Minigrid lava environment.}
  \label{fig_lava}
\end{figure}%

\section{Results}

\subsection{Performance of pruned policies}
\label{sec:prunedpol}
 The precise ranking of decisions according to their importance for the reward is intractable for all 
but very simple policies. We
use the performance of 
\emph{pruned policies} as a \emph{proxy} for the quality of the ranking computed by our algorithm.
In pruned policies, all but the top-ranked actions are replaced by randomly sampled actions.
For a given $r$ (a fraction or a percentage), we denote
by $\mathit{rank}[r]$ the subset of $r$ top-ranked states.
We denote by~$\pi^r$ the pruned policy obtained by \emph{pruning}
all but the top-$r$ ranked states. That is, an execution of~$\pi^r$ retains actions in the $r$ fraction
of the most important states from the original policy $\pi$ and replaces the rest
by random actions. The states that are in
$\mathit{rank}[r]$ are called the \emph{original states}.
We measure the performance of the pruned policies for increasing values of $r$ relative to
the performance of the original policy $\pi$.
To evaluate the quality of each ranking method, 
we measure how quickly we are able to recover the
performance of the original policy~$\pi$
as we reduce the set of pruned states. We start with $r=0$,
and evaluate the performance of $\pi^r$ 
for increasing values of $r$. 

We consider how performance is recovered as 
the percentage of decisions drawn from the original policy increases,
and as the percentage of steps 
in which the original policy is used increases
(note that for the former, we always prune all of 
the states not encountered in the test 
suite, even at 100\%).
These two ways of reporting performance
differ when, for example, we rank highly
states that are important if visited but rare. Even if 
many of theses states are not pruned, 
the policy would still more often take a random 
action, despite fewer states being pruned overall.

\begin{figure}
        \centering
        \includegraphics[width=0.47\linewidth]{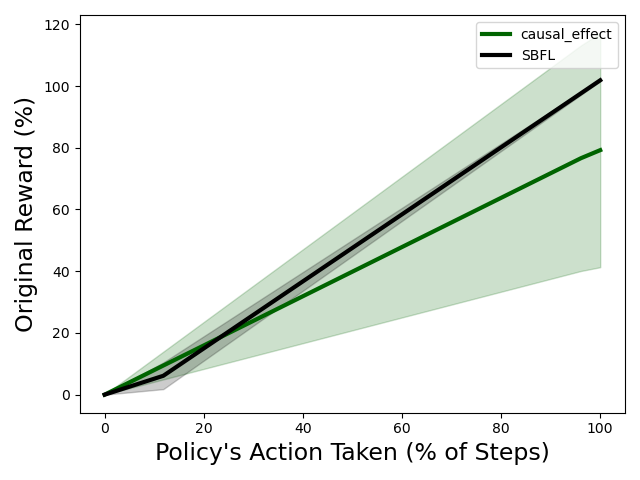}\hfill%
        \includegraphics[width=0.47\linewidth]{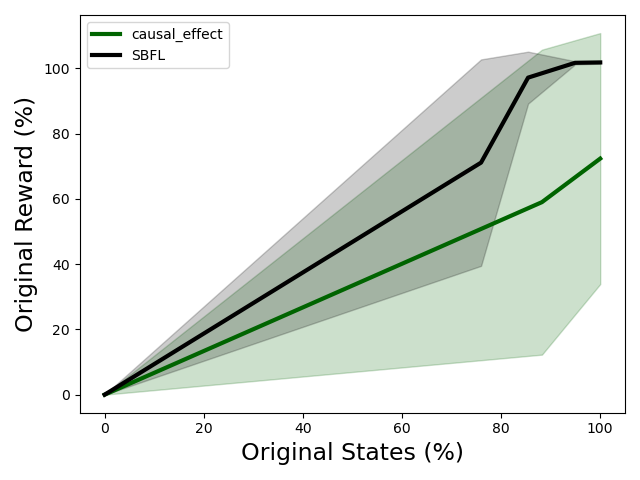}
        \caption{Comparing causal ranking and SBFL-based ranking in the minigrid lava environment based on the percentage of the original reward recovered based on a pruned policy (left). The percentage of original states occupied versus reward recovered under the pruned policy (right). Lines (shaded regions) indicate the mean (standard error) across 20 experimental runs.}
        \label{fig_results}
\end{figure}

We applied our analysis to twenty challenging minigrid environments in which the agent has to reach the goal and avoid falling into the lava (see an example in Figure~\ref{fig_lava}). Our preliminary results (see 
Fig.~\ref{fig_results}) show that causality-based policy ranking provides a level of performance comparable to that of SBFL-based policy ranking in~\cite{PCSK20} (see figure captions for further details). 

\FloatBarrier
\section{Discussion}

We presented some initial results on policy simplification via ranking policy decisions based on counterfactual reasoning. The performance of the resulting policies was comparable to previously established baselines for this problem \citep{PCSK20}. Despite the lack of an overall performance improvement, we suggest that causality-based policy ranking provides a more direct and interpretable methodology. Furthermore, we argue that this work establishes the potential for causality-based policy ranking with a rich potential for future work. Beyond causal effects measures, more sophisticated causal inference algorithms such as the PC and FCI algorithms may be deployed in the service of policy ranking \citep{Pearl2009}. Such methodologies may, for example, be able to adjust for complex interactions between decisions across timepoints. This could account for higher-order dependencies in policy decisions whereby combinations of actions (or action avoidances) in different states may be identified as contributing significantly to reward maximization. Additionally, we hypothesize that recent developments in score-based causal analysis \citep{Zheng2018a} may allow our policy ranking method to be extended to continuous control problems.

\bibliographystyle{abbrvnat}
\bibliography{references}

\end{document}